\documentclass[a4paper, 10pt, conference]{ieeeconf}

\IEEEoverridecommandlockouts 
\usepackage{times}
\usepackage{comment}
\usepackage{soul}
\usepackage{multicol}
\usepackage{multirow}
\usepackage[bookmarks=true, hidelinks]{hyperref}
\usepackage{adjustbox}
\usepackage{tikz}
\usetikzlibrary{shapes,arrows,positioning,calc, shapes.geometric}
\tikzstyle{decision} = [diamond, minimum width=3cm, minimum height=1cm, text centered, draw=black]
\usepackage{mathptmx} 
\usepackage{amsmath} 
\usepackage{amssymb}
\usepackage{graphicx}
\usepackage{textcomp}
\usepackage{algorithm,algpseudocode}%
\usepackage{cite}
\usepackage{balance}
\usepackage{mathptmx}
\usepackage{subcaption}
\usepackage{lipsum}
\let\labelindent\relax
\usepackage{enumitem}
\usepackage{graphicx}
\usepackage[english]{babel}
\usepackage{array}
\usepackage{mathtools}
\usepackage{stfloats} 
\usepackage{float}
\usepackage{tabularx}

\usepackage{color, colortbl}
\definecolor{LightCyan}{rgb}{0.88,1,1}
\definecolor{LightRed}{rgb}{1.,0.8,0.8}

\algnewcommand{\Inputs}[1]{
  \State \textbf{Inputs:}
  \Statex \hspace*{\algorithmicindent}\parbox[t]{.8\linewidth}{\raggedright #1}
}
\algnewcommand{\Initialize}[1]{
  \vspace{5pt}\State \textbf{Initialize:}
  \Statex \hspace*{\algorithmicindent}\parbox[t]{1\linewidth}{\raggedright #1\vspace{5pt}}
}

\makeatletter
\newcommand{\thickhline}{%
    \noalign {\ifnum 0=`}\fi \hrule height 1.5pt
    \futurelet \reserved@a \@xhline
}
\newcolumntype{"}{@{\hskip\tabcolsep\vrule width 1.5pt\hskip\tabcolsep}}
\newcolumntype{^}{@{\hskip\tabcolsep\vrule width 2pt\hskip\tabcolsep}}

\makeatother

\newcommand\figref{Fig.~\ref}

\tikzset{
block/.style = {draw, fill=white, rectangle, minimum height=3em, minimum width=3em},
tmp/.style  = {coordinate}, 
sum/.style= {draw, fill=white, circle, node distance=1cm},
input/.style = {coordinate},
output/.style= {coordinate},
pinstyle/.style = {pin edge={to-,thin,black}
}
}

\title{\LARGE \bf
Benchmarking local motion planners 
\\ for navigation of mobile manipulators
}

\author{Sevag Tafnakaji$^{\dagger}$ \hspace{10pt} Hadi Hajieghrary$^{\dagger}$ \hspace{10pt} Quentin Teixeira$^{\dagger}$  \hspace{10pt} Yasemin Bekiroglu$^{\dagger, *}$
\thanks{
$^{\dagger}$Department of Electrical Engineering, Automatic Control Research Group, Chalmers University of Technology, Göteborg, Sweden. 
$^*$University College London, Statistical Machine Learning Group, AI Center, London, UK.
\tt{\small{Email: sevag@student.chalmers.se}
}}}

\begin{document}
\maketitle

\thispagestyle{empty}
\pagestyle{empty}

\begin{abstract}
                                                                                                                                            
There are various trajectory planners for mobile manipulators. It is often challenging to compare their performance under similar circumstances due to differences in hardware, dissimilarity of tasks and objectives, as well as uncertainties in measurements and operating environments. In this paper, we propose a simulation framework to evaluate the performance of the local trajectory planners to generate smooth, and dynamically and kinematically feasible trajectories for mobile manipulators in the same environment. We focus on local planners as they are key components that provide smooth trajectories while carrying a load, react to dynamic obstacles, and avoid collisions. We evaluate two prominent local trajectory planners, Dynamic-Window Approach (DWA) and Time Elastic Band (TEB) using the metrics that we introduce. Moreover, our software solution is applicable to any other local planners used in the Robot Operating System (ROS) framework, without additional programming effort.

\end{abstract}

\section{INTRODUCTION}

Motion planning is a key component 
for mobile robotic applications \cite{marder-eppstein_office_2010, hidalgo-paniagua_solving_2017, zhang_multilevel_2019, wang_eb-rrt_2020}. In order to reduce the complexity, motion planning is often divided into two logical parts: global and local planning \cite{kudriashov_introduction_2020}. A global planner searches for the overall path toward the goal on the static map of the environment. The local planner uses the sensory information of the robot to perceive the current state of the robot's immediate surroundings and generates feasible trajectories to be followed by the robot. Avoiding any dynamic or static obstacles that may or may not have been included in the given global map is the responsibility of the local planner. The local planner fuses data from sensors such as laser scanners, ultra-sound sensors, and various cameras to predict the state of the immediate surroundings of itself, and plan a dynamically and kinematically feasible trajectory for the mobile robot to follow. Therefore, the performance of the local planner will directly affect the overall success outcome of a mobile robot's task. In the case of a mobile manipulator, this matter is of paramount importance, since any rugged motion of the mobile base may translate into catastrophic trembling of the payload grasped by the manipulator. 

\begin{figure}[!ht]
     \centering
    \includegraphics[width=1\linewidth]{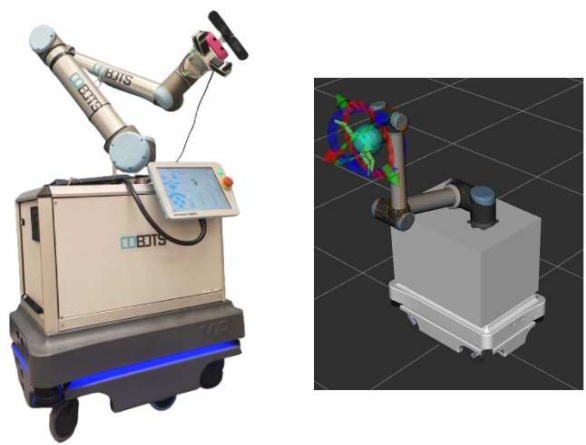}
    \caption{(Left) A mobile manipulator, including a Mobile Industrial Robots MiR mobile base, a Universal Robot UR10 robotic arm, and OnRobot RG6 gripper. (Right) The corresponding robot model 
    shown in Rviz (3D visualization tool from ROS).}
    \label{fig:mobile_manipulator}
\end{figure}

In this paper, we present a framework to evaluate the performance of a local planner and use the framework to benchmark two predominant methods, the latest implementation of Dynamic-Window Approach (DWA), called DWB, \cite{fox_dynamic_1997}, and Time Elastic Band (TEB) \cite{rosmann_integrated_2017, roesmann_trajectory_2012}. We evaluate their performance 
in terms of generating smooth feasible trajectories under various conditions. Nevertheless, many of the presented methodologies and provided machinery can be easily used to benchmark other local planners. The result of the benchmark presented in this document aims to determine how effectively a planner navigates an industrial mobile manipulator moving a heavy load. We introduce an extensive set of metrics, and hypothesize that these measures will paint a clear picture about the performance of a mobile manipulator. Then we test our hypothesis by inspecting the performance of a realistically simulated industrial mobile manipulator in different environments. 

The Dynamic-Window Approach to planning the trajectory for a mobile robot constructs the velocity space around the robot, i.e., a space which is accessible with a span of the possible linear and angular velocities. Within this space a small window around the current pose of the robot is chosen. Any of the velocity profiles which may lead the robot to collision with an obstacle is removed. The remaining set of velocities that are deemed safe is sampled, and the samples are simulated to determine where they would drive the robot if they were applied. The velocity commands that offer the best result are chosen as the control signal and then sent to the robot. The performance is usually measured based on the advance the robot will make along the global path.

The DWB is a sampling-based algorithm. In contrast, the Time Elastic Band approach (TEB) is a potential-field optimization-based algorithm. The TEB approach is designed to deform and optimize the global trajectory, which is initially coarsely planned and is purely geometric. The deformation of the global trajectory is guided by the presence of potential fields which are to design to shorten the path and obtain the shortest path, while maintaining a separation between the path and obstacles \cite{Quinlan1993}. 

The primary task of a local planner is to turn a path generated by the global planner into a dynamically feasible trajectory, which can be followed by the robot. The similarly important task of the local planner is to re-evaluate the path generated by the global planner, and determine if that path is still feasible. The initially proposed path might no longer be feasible due to several factors: the map uploaded to the robot may not be valid anymore, since some changes may have occurred in the environment; or a unexpected obstacle is found on the path of the robot. In both cases, it is the responsibility of the local planner to generate a feasible and safe trajectory for the robot to follow. In this paper, we are testing the performance of the local planner. However, it would be quite easy to deploy this framework to benchmark any dynamic obstacle avoidance algorithm in the Robot Operating System (ROS) Framework.  

\begin{figure}[t]
    \centering
    \begin{subfigure}[b]{0.44\textwidth}
       \centering
        \includegraphics[width=0.9\textwidth]{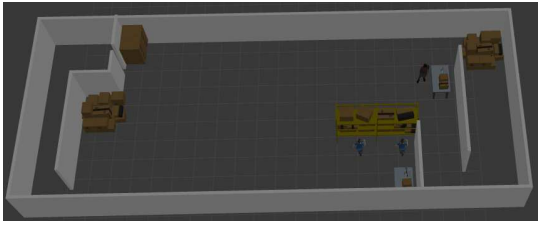}
        \caption{Playground}
        \vspace{10pt}
        \label{fig:playground_dynamic}
    \end{subfigure}
    \begin{subfigure}[b]{0.44\textwidth}
       \centering
        \includegraphics[width=0.9\textwidth]{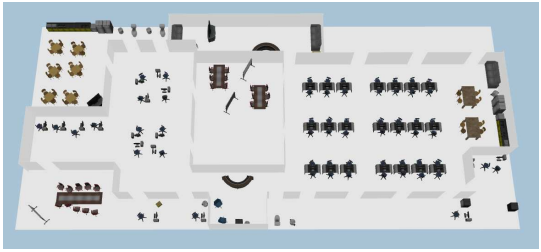}
        \caption{Office}
        \vspace{10pt}
        \label{fig:office_dynamic}
    \end{subfigure}
    \begin{subfigure}[b]{0.44\textwidth}
       \centering
        \includegraphics[width=0.9\textwidth]{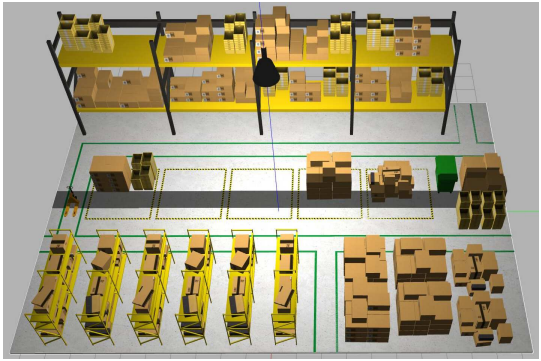}
        \caption{Warehouse}
        \label{fig:warehouse_dynamic}
    \end{subfigure}
    \caption{Simulated Gazebo worlds with/without dynamic obstacles, used to benchmark the TEB and DWB algorithms.}
    \label{fig:worlds}
    \vspace{-15pt}
\end{figure}

The proposed benchmarking framework is implemented in ROS. We run it for a simulated mobile manipulator (see Fig.\ref{fig:mobile_manipulator}) inside the Gazebo simulation. In this paper, we present the results of the benchmarking of DWB and TEB in guiding a mobile manipulator using three Gazebo worlds: playground, warehouse, and office environment, see \figref{fig:worlds}. In each case, we benchmark the performance of the local planner for two scenarios: first, the complete map of the environment is available for the robot with high fidelity. Second, some objects in the environment are not included in the map the robot possesses. In all of these tests, the performance of the local planners is measured using the metrics we have defined to quantify the smoothness of the planned path, the smoothness of the motion of the manipulator's end-effector, and the discrepancy between the planned global path and the actual one. Also, we compare the algorithms based on overall performance metrics such as total time, distance traveled, and final position accuracy. 

In \cite{pittner_systematic_2018} and \cite{cybulski_accuracy_2019}, motion planning algorithms such as DWB, TEB, and E-band are tested using small mobile robots in both simulation and real experiments. The general consensus is that DWB performs the fastest, E-band is the most accurate local planner, and TEB plans motions that lead the robot to the goal slightly faster than the other two algorithms. The authors also report that TEB would react the fastest to a dynamic obstacle, while DWB sometimes stops to calculate its subsequent route. In \cite{marin-plaza_global_2018} and \cite{naotunna_comparison_2020}, similar results are reported with a heavy differential-driven robot carrying a Baxter manipulator. These results indicate that in general, TEB performs better than the baseline methods 
at obstacle avoidance in simulation and real experiments. 

In order to improve navigation benchmarking efforts,  different standards are proposed, \cite{hsieh_experimental_2016}, \cite{wen_mrpb_2021}. In \cite{hsieh_experimental_2016}, the authors introduce the standard of worlds with different types of obstacles. They also propose adding variations in how the dynamic obstacles are set up (from simpler to more complicated environments) and perform experiments in real and simulation environments. 
There are also studies, \cite{wen_mrpb_2021} focusing more on crafting standard metrics that can meaningfully benchmark the navigation algorithms using standard worlds/environments.

Metrics used in evaluations are key to analyze the performance of the algorithms. In \cite{wen_mrpb_2021}, the authors have found that TEB generates much smoother paths when using a metric they introduced. This metric constitutes the sum of the difference between two position measurements when the mobile base is traveling along a path. However, this method may fail to discern how smooth the generated path is. 
In \cite{guillen_ruiz_measuring_2020}, the authors simulate a mobile robot in different scenarios to measure the generated path's smoothness. 
They analyze energy consumption in a real robot for trajectories using different smoothness factors.
They also analyze social acceptance for different smoothness factors by presenting different simulated situations to different people. Their results indicate that in general smoother paths decrease energy consumption and increase acceptability, as long as important factors such as distance to people are fulfilled.

Our proposed framework is different from the previous approaches in the following aspects: we propose a framework to evaluate the performance of local trajectory planners to generate dynamically and kinematically feasible trajectories for a mobile manipulator carrying a load; and, we concentrate on the smoothness of the motion for the load carried by the manipulator. In addition, our benchmarking measures indicate, in return, how the motion of the manipulator affect the maneuver of the mobile base. 

Studying the stability of the local planners in this context reveals the necessary precautions that should be taken while designing a mobile manipulator to transport sensitive material, as mobile manipulators are more complex and have more noise sources than the standalone mobile robot. There have been many benchmarks and evaluations on local planners for using mobile robots, but not many that focus on applications of mobile manipulators. This paper focuses on the performance of local planners for the specific use cases of a mobile manipulator rather than a more simple mobile robot. With the added complexity of the arm attached to the base, there are more components and aspects that the local planner can affect. One of these, which is the paper's primary focus, is how smoothly the mobile manipulator can carry loads. The ability of the robot to transport a load can be hindered due to the poor performance of the local planner. If the abilities of a mobile manipulator is not quantified systematically, the user will limit its use due to, mainly, safety reasons, and leave much of its economic potential unrealized. The proposed framework to benchmark a local planner in particular, and a mobile manipulator in general, is intended to help devise an optimal navigation stack to fully exploit a mobile manipulator's functionalities 
in an industrial setting.

\section{PROPOSED METHOD}
\label{section:method}

In this paper, we benchmark the local planning algorithms for mobile manipulators in terms of 1) the total time taken to traverse from the origin to the goal; 2) the total distance traveled to reach the goal ; 3) the accuracy of the position of the end-effector; 4) the capability of the algorithm to avoid the obstacles; and 5) the smoothness of the planned path. 


The procedure is repeatable, i.e., it is iterated for the desired number of times using the same test. The user can choose and modify different parameters of the planner, as well as the simulated system. During each trial, all the necessary measurements are collected to create the set of proposed metrics to analyze the performance of the local planner:

\begin{enumerate}[label=\textit{\Alph*)}]
    \item \textit{Path smoothness:}
    \label{subsection:path_smoothness}
    As the robot travels along the global path, a measurement of its pose is sampled and stored. With enough measurements, i.e., $N$, the taken path is retraced by simply treating the pose as vectors and finding how smoothly the robot travels between measurements, i.e., $\Delta\mathbf{x}_i = x_i - x_{i-1}, i = 2, \dots, N-1$, where $x_i$ is the measurement of the position of the robot in  Cartesian coordinates. We can find the angle of a path segment as the angle between two succeeding positions and present the measure of path smoothness, $p_s$, as:
    \begin{equation}
        p_s = \sum\alpha_i,
    \end{equation}
   where, $\alpha_i$ is the angle between the vectors $\mathbf{x}_i$ and $\mathbf{x}_{i+1}$. The desired smooth path will be the one that there is only a slight angle between each two consequent vectors; preferably, the angle would be zero; hence, the optimal value for this metric would be as small as possible. Another approach for this metric could be to take the inverse of the sum of all angles, which would make it desirable to maximize this value.
    
    \item \textit{End-effector velocity smoothness:}
    This measures the discrepancy between the actual configuration of the robot - more specifically, the pose of the end-effector, with the expected pose of the end-effector based on the designed trajectory for the mobile base. We define the difference between the expected position of the end-effector, $\mathbf{p}_{exp}$, and the measured position of the end-effector,$\mathbf{p}_{act}$, as:
    \begin{equation}
        \mathbf{e} = \mathbf{p}_{exp} - \mathbf{p}_{act}.
    \end{equation}
    Once the error is calculated, we integrate the absolute value of the error over the entire traveled path. This results in a quantity which quantifies the stability of the load being transported by the mobile manipulator:
    \begin{equation}
        \mathbf{p}_e = \begin{bmatrix}\ p_{e_x} \ \\ \ p_{e_y} \ \\ \ p_{e_z} \end{bmatrix} = \int_T\ \left| \mathbf{e}\ \right| dt = \begin{bmatrix}\ \int_T\ \left| e_x \right|\ dt \ \\ \ \int_T\ \left|e_y\right|\ dt \ \\ \ \int_T\ \left|e_z\right|\ dt \ \end{bmatrix}.
    \end{equation}
    $T \in [t_{start},t_{final}]$ is the time over which we integrate and $t_f$ is the time it takes for the robot to complete its travel along the path. The expected path of the end-effector can be found in many ways, depending on whether the arm is moving or not. If the arm is moving, then by using the joint angles and the corresponding forward kinematic equations one can find the expected pose. If the arm is not moving and is in a known pose, then by using the simulation software's own physics engine measurements for the base and applying a simple translation, we can find the expected pose. The actual pose would be measured directly from the simulation. In a real scenario, the actual pose may be measured with an external sensor or with a camera mounted on the manipulator.
    
    \item \textit{Distance travelled:}
    This metric is the sum of Euclidean distances between the sampled position of the end-effector:
    \begin{equation}
        d_{travelled} = \sum^{N-1}_{i=2}\vert\vert\Delta\mathbf{x}_i\vert\vert
    \end{equation}

    \item \textit{Integrated difference between global and local path:}
    As the travelled path deviates from the global path, the area between them can be approximated by taking the distance between the paths:
    \begin{equation*}
        d_{between} = \sum^{\text{min}(N,n)}_{i=1} \vert\vert \mathbf{x}_i - \mathbf{X}_i\vert\vert^2
    \end{equation*}
    where $n$ is the number of way-points on the planned path, $\Delta\mathbf{X}_i$ is the position of the robot at the planned path (taken from the global planner), and $\Delta\mathbf{x}_i$ is the actual position of the robot (taken from the local planner). The area between the global and local paths can be approximated as:
    \begin{equation}
        A_{\text{between}} = \dfrac{d_{between}\cdot d_{travelled}}{\text{min}(N,n)}
    \end{equation}

    \item \textit{Final position accuracy:}
    The final position accuracy is measured when the navigation is done, and it is simply the Euclidean distance between the goal position, $\mathbf{p}_{goal}$,  and the final position of the robot, $\mathbf{p}_{final}$:
    \begin{equation}
        p_{acc} = \vert\vert\mathbf{p}_{final} - \mathbf{p}_{goal}\vert\vert^2
    \end{equation}
    It is important to note that this metric does not account for the orientation difference between the final pose and the goal pose of the robot. 
    
    \item \textit{total time taken:}
    Once the ROS navigation stack receives a goal position, a timer is started at time $t_{start}$ and stops once the robot reaches the goal at time $t_{final}$:
    \begin{equation}
        T_{taken} = t_{start} - t_{end}
    \end{equation}
    The total time consists of two essential parts: the time taken for computations and the time taken to travel along the computed path. 
\end{enumerate}

It is important to note that $x_i$ is measured from the centre of the mobile base MiR, whereas $\mathbf{p}_{exp}$ and $\mathbf{p}_{act}$ is the position of the end-effector (expected and actual, respectively).

\section{RESULTS}
\label{section:simResults}

\begin{figure}
	\centering
    \begin{subfigure}[b]{0.49\linewidth}
       \centering
       \includegraphics[width=1\linewidth]{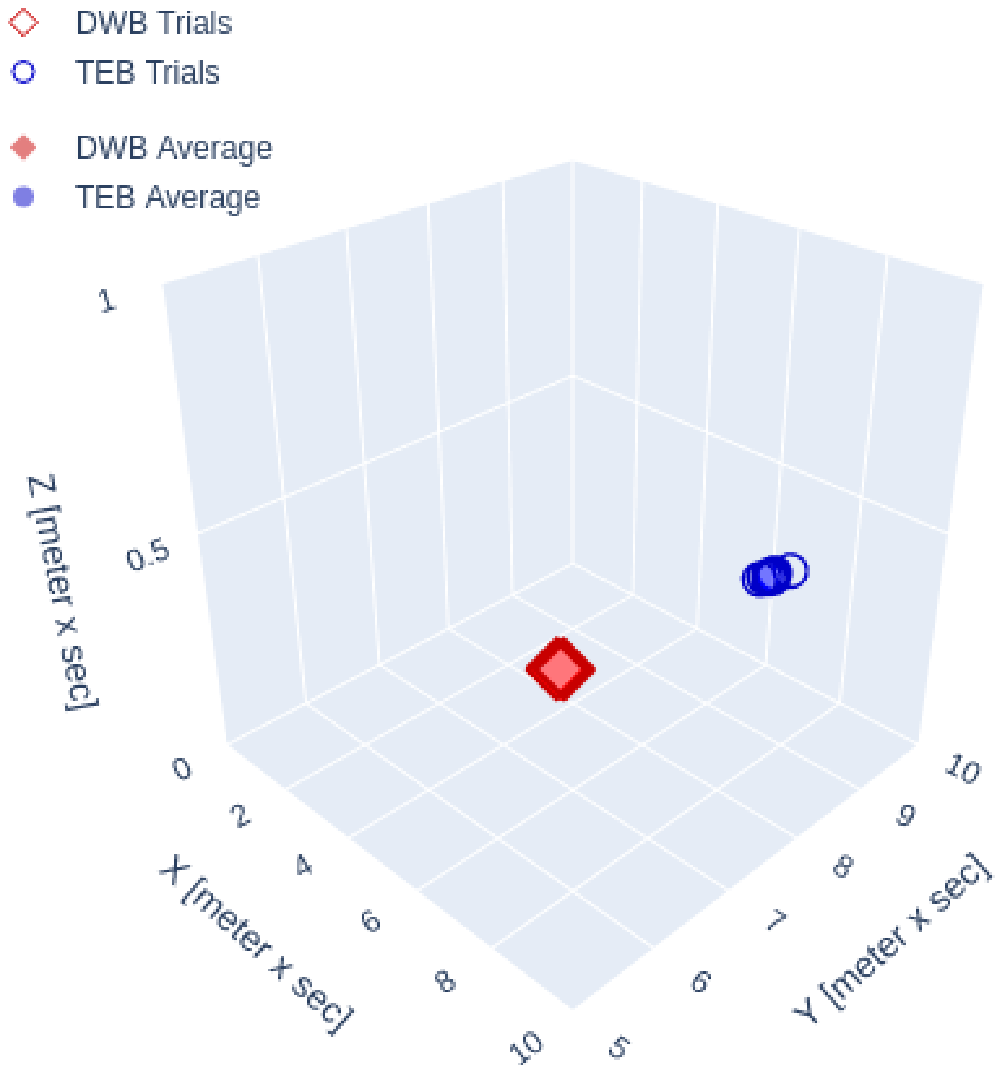}
       \caption{Static Playground}
    \end{subfigure}
    \begin{subfigure}[b]{0.49\linewidth}
       \centering
       \includegraphics[width=1\linewidth]{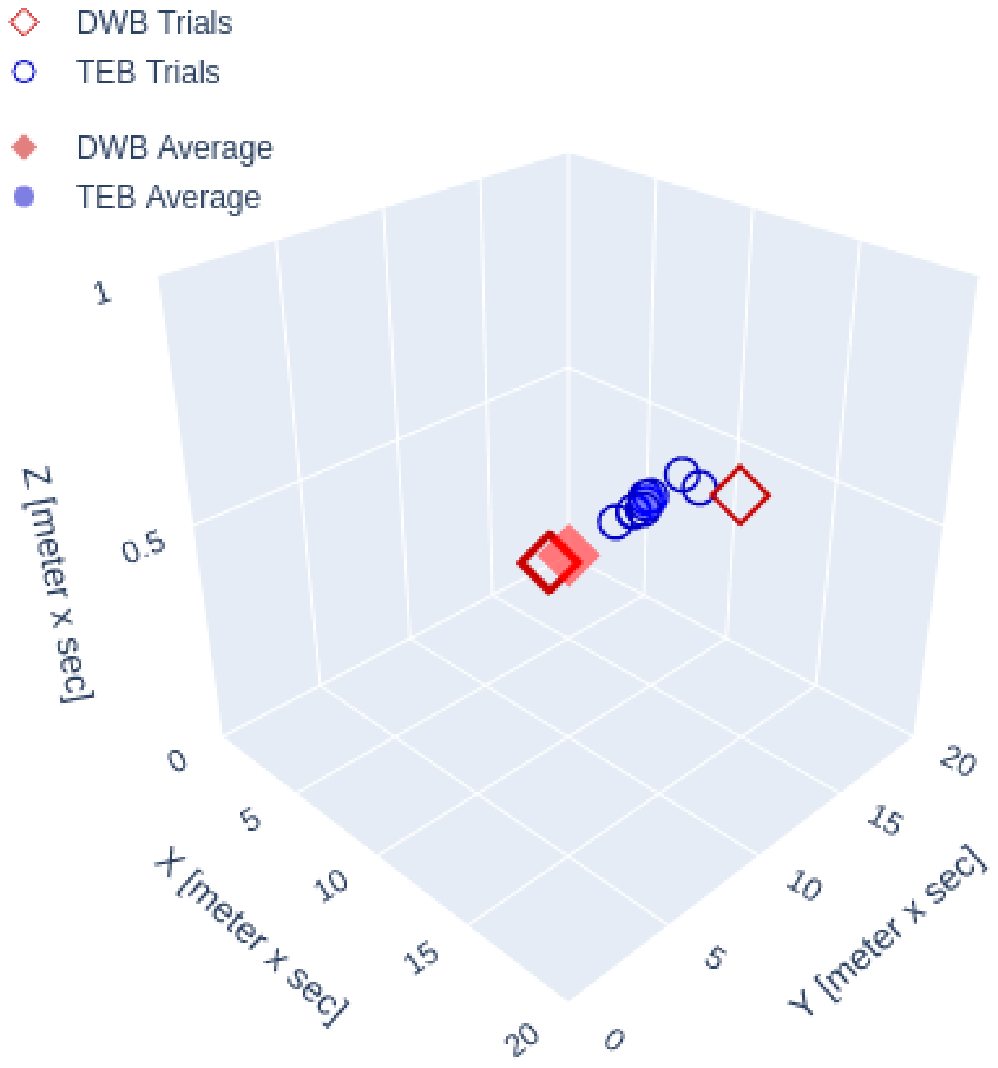}
       \caption{Dynamic Playground}
    \end{subfigure}
    \begin{subfigure}[b]{0.49\linewidth}
       \centering
       \includegraphics[width=1\linewidth]{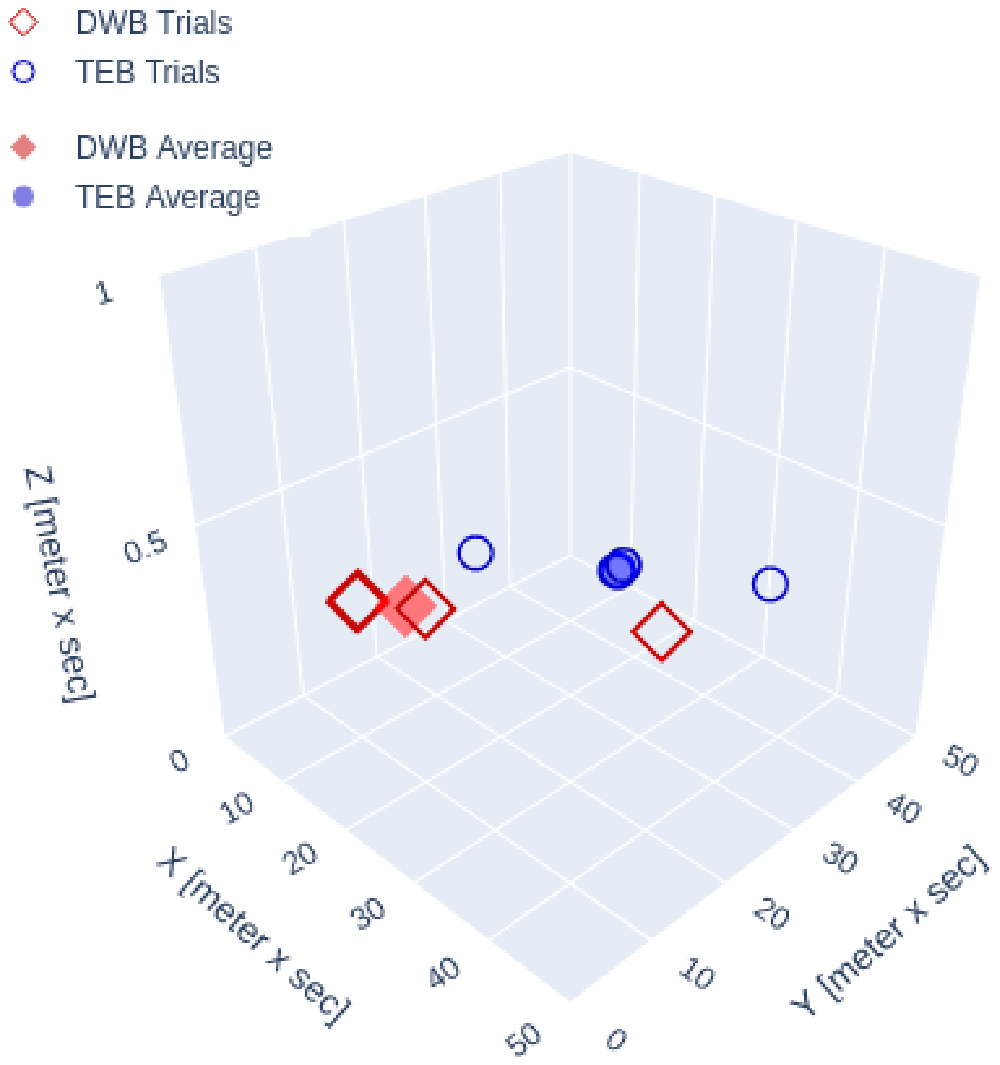}
       \caption{Static Office}
    \end{subfigure}
    \begin{subfigure}[b]{0.49\linewidth}
       \centering
       \includegraphics[width=1\linewidth]{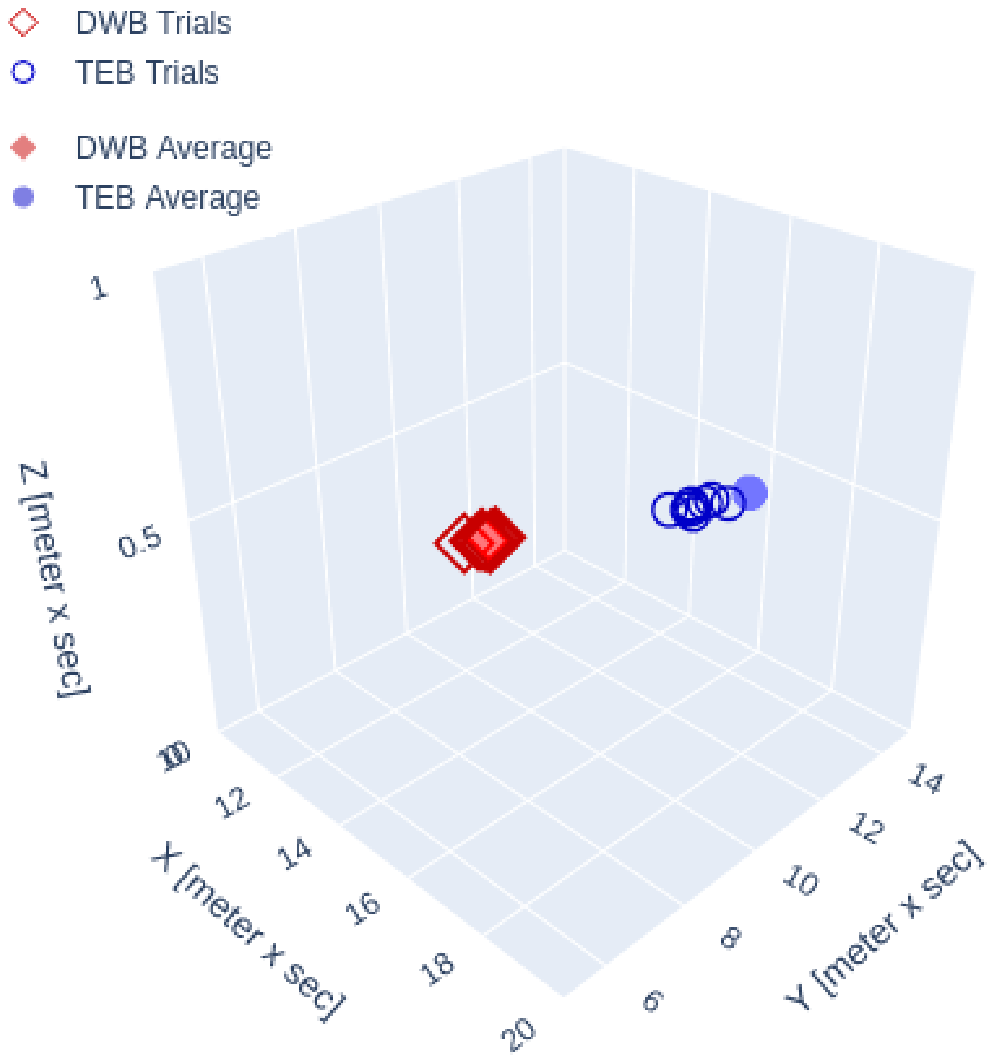}
       \caption{Dynamic Office}
    \end{subfigure}
    \begin{subfigure}[b]{0.49\linewidth}
       \centering
       \includegraphics[width=1\linewidth]{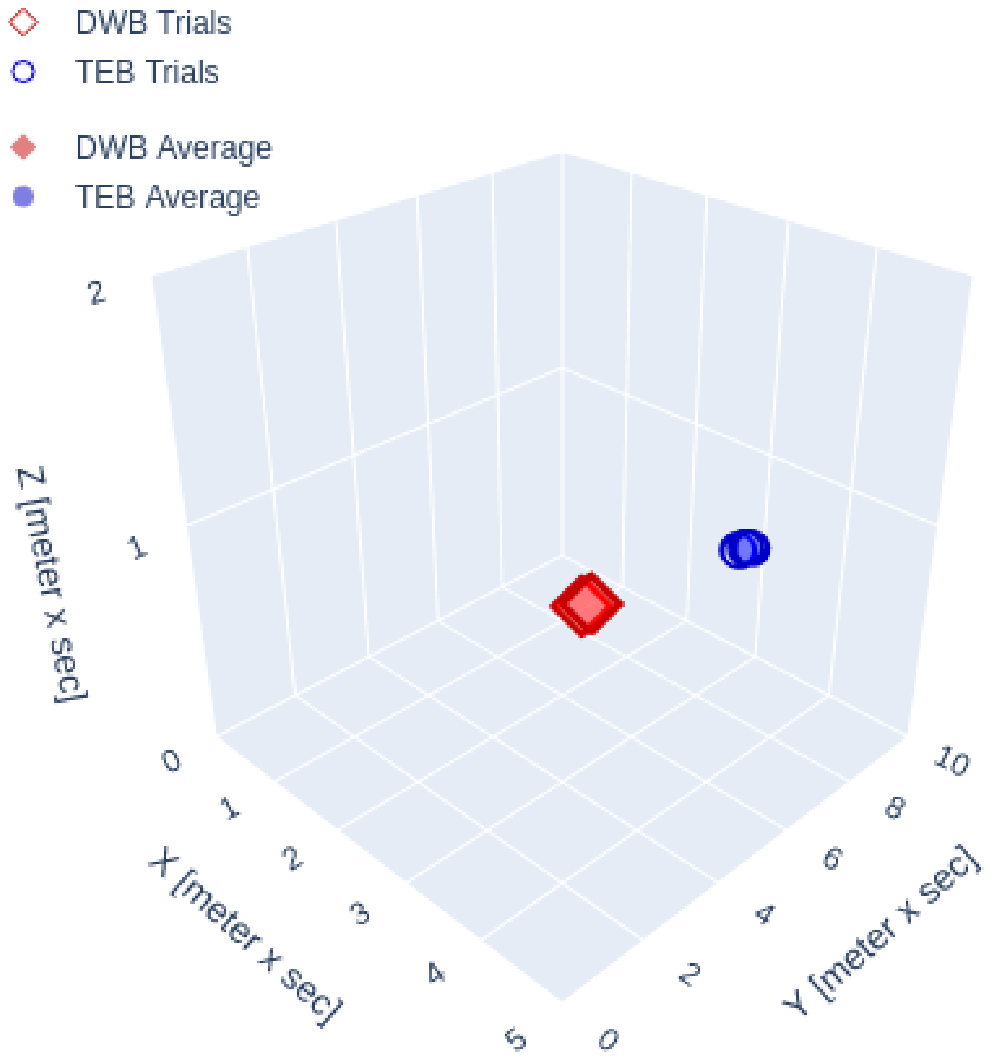}
       \caption{Static Warehouse}
    \end{subfigure}
    \begin{subfigure}[b]{0.49\linewidth}
       \centering
       \includegraphics[width=1\linewidth]{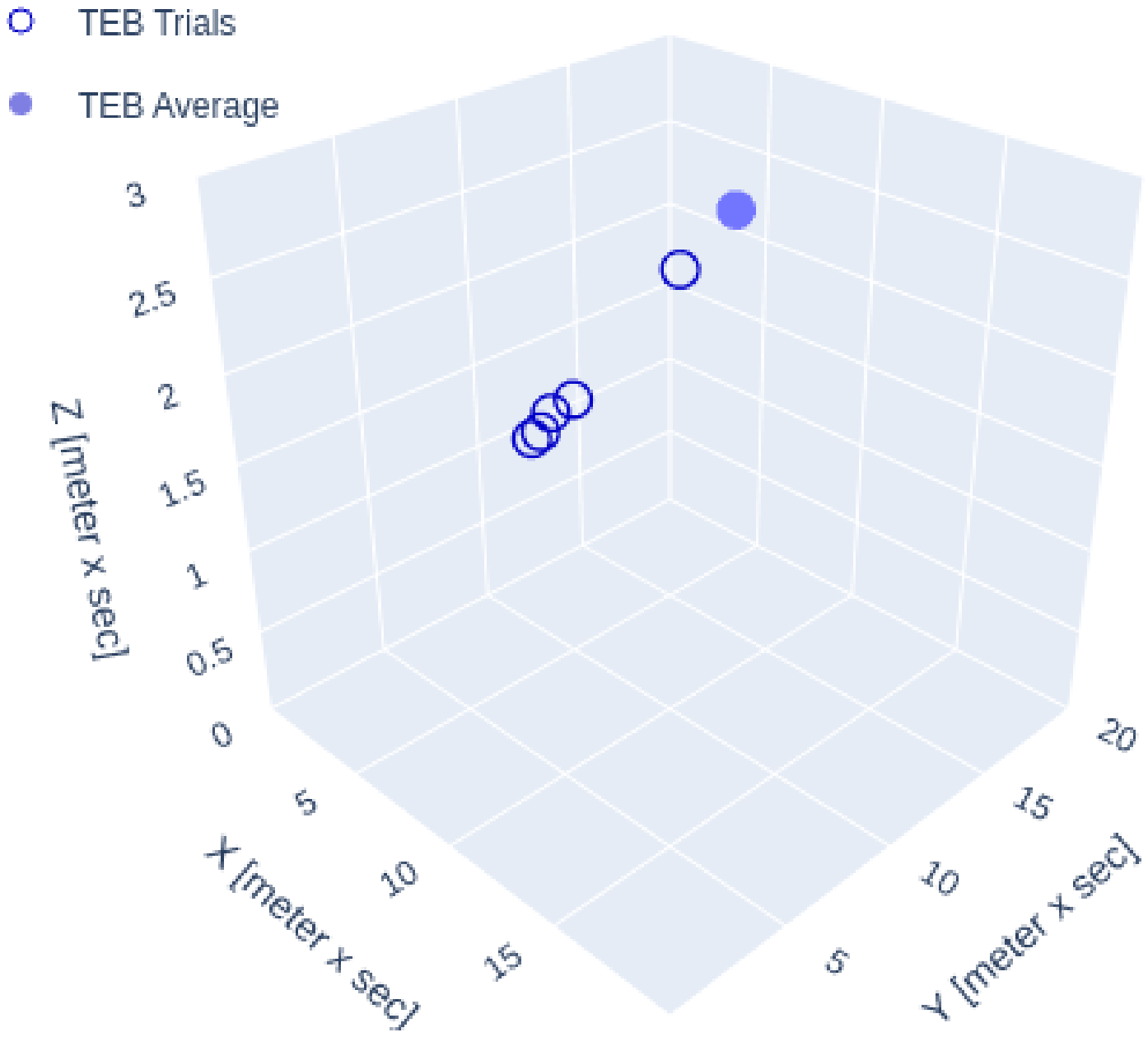}
       \caption{Dynamic Warehouse}
    \end{subfigure}
    \caption{Stability metric, $\mathbf{p}_e$, indicating the difference between the expected trajectory of the end-effector and its real trajectory while the mobile manipulator is navigating through the environments using TEB and DWB local planners. TEB results in more smooth motions for the end-effector compared to DWB, particularly, in dynamic environments. In the dynamic warehouse DWB was not able to plan any trajectory to avoid the obstacle.}
     \label{fig:ee_smoothness}
\end{figure}

\begin{figure}
	\centering
    \begin{subfigure}[b]{0.49\linewidth}
      \centering
      \includegraphics[width=1\linewidth]{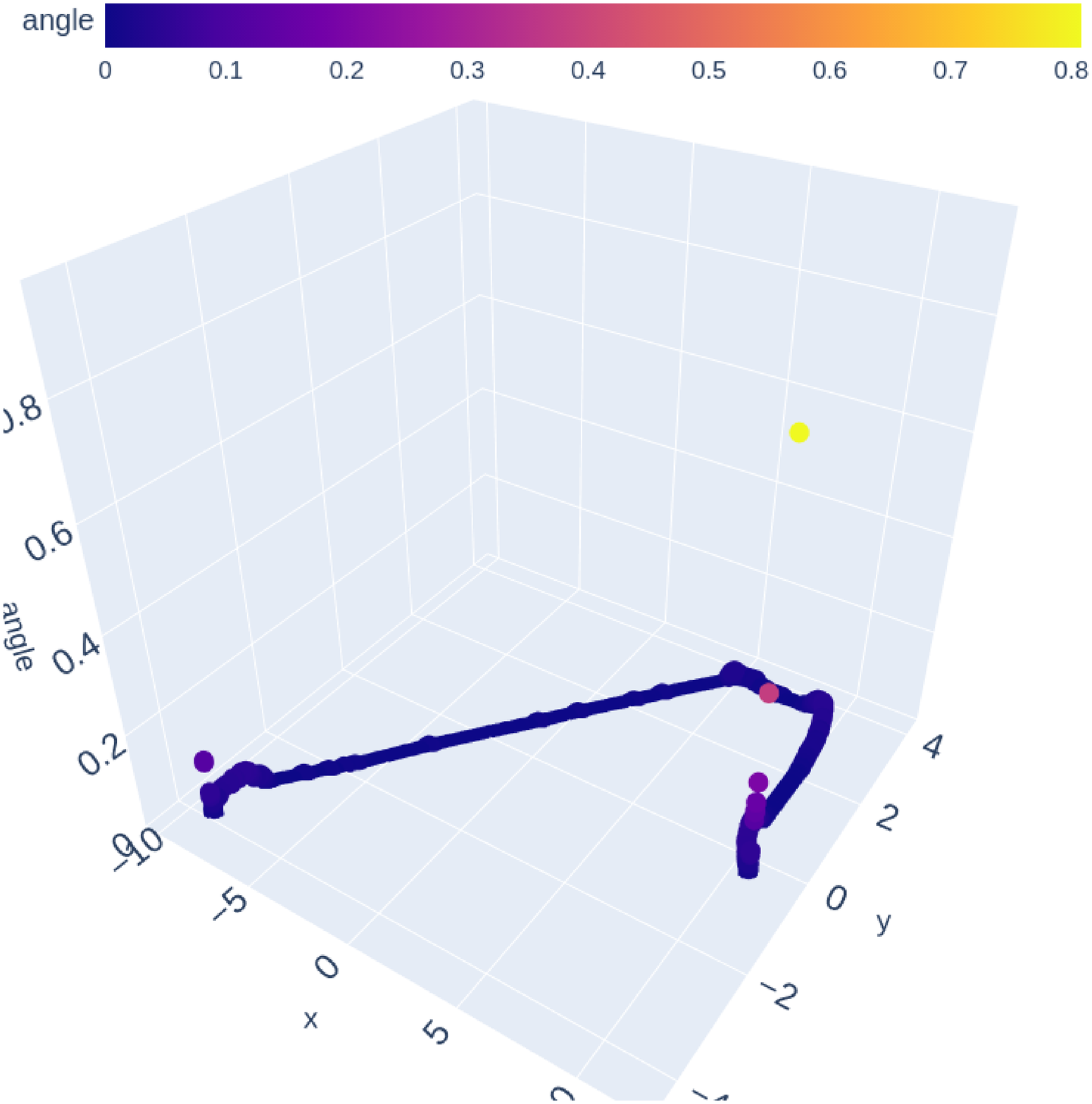}
      \caption{Static - DWB}
    \end{subfigure}
    \begin{subfigure}[b]{0.49\linewidth}
      \centering
      \includegraphics[width=1\linewidth]{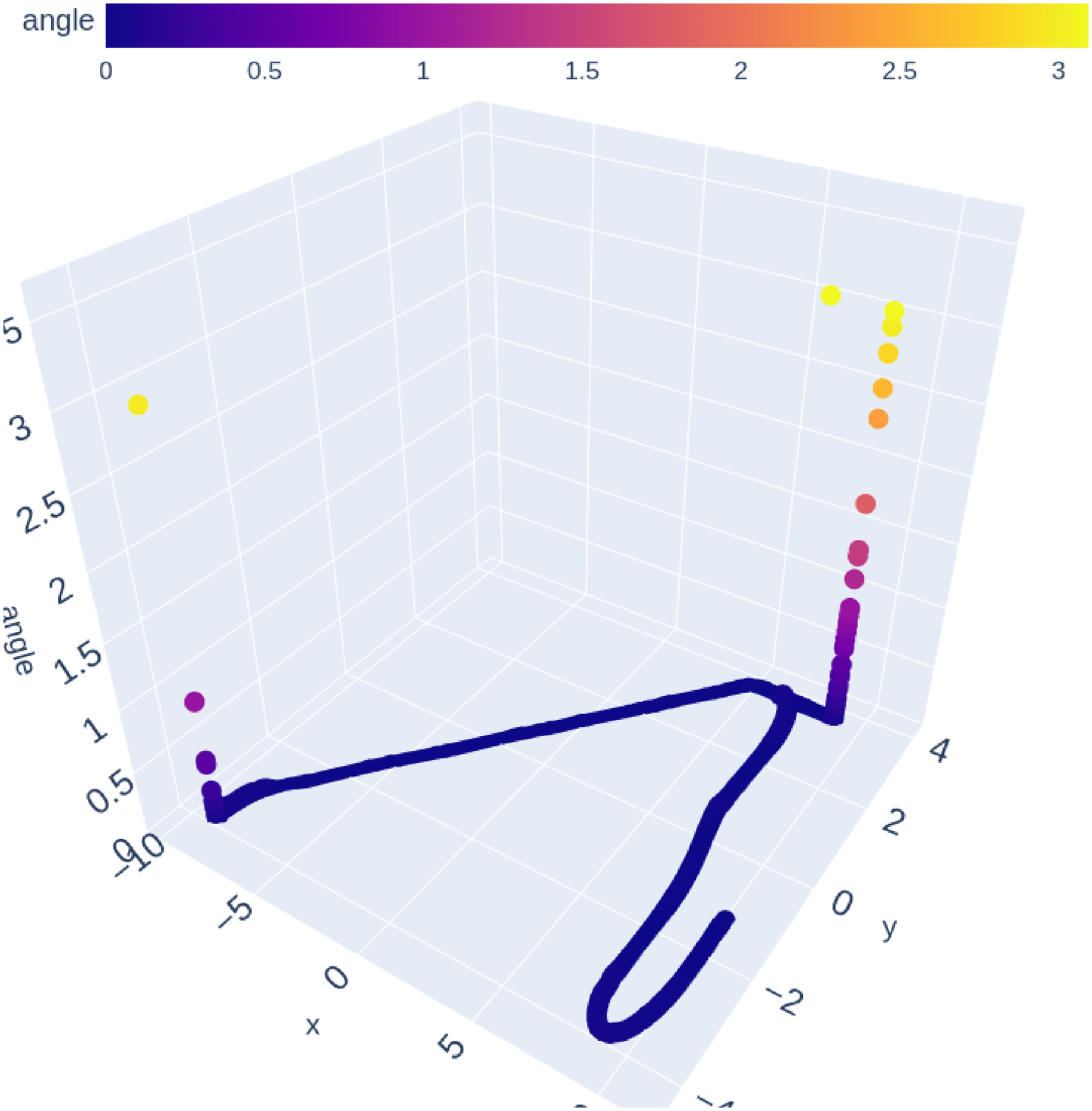}
      \caption{Dynamic - DWB}
    \end{subfigure}
    \begin{subfigure}[b]{0.49\linewidth}
      \centering
      \includegraphics[width=1\linewidth]{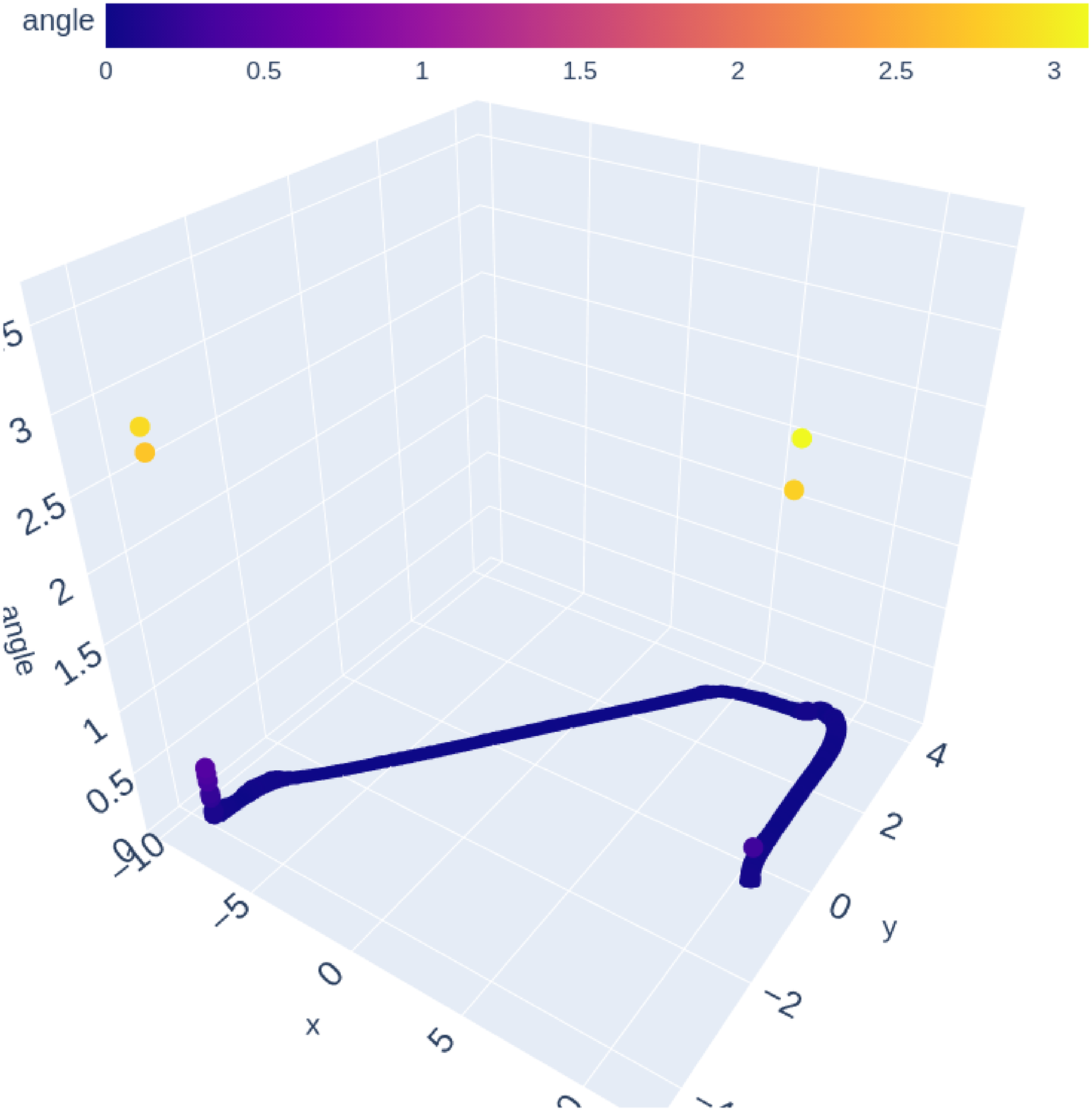}
      \caption{Static - TEB}
    \end{subfigure}
    \begin{subfigure}[b]{0.49\linewidth}
      \centering
      \includegraphics[width=1\linewidth]{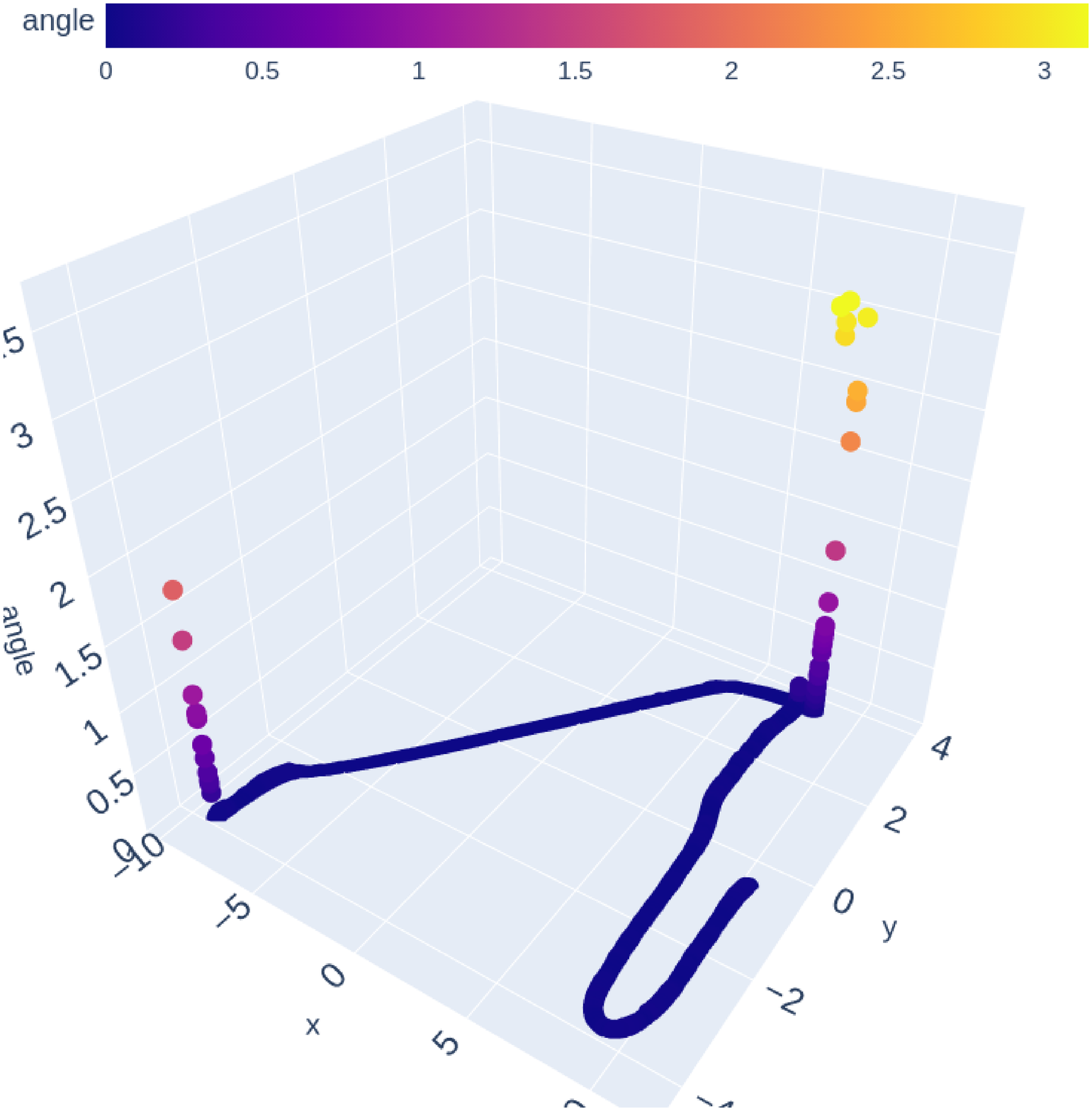}
      \caption{Dynamic - TEB}
    \end{subfigure}
    \caption{Plot of the travelled path (x,y) along with the smoothness at that point as the z-axis, showing the angle between the current direction and the new direction. This example shows the results from one trial using the playground world, and how the addition of obstacles affects the smoothness. The added obstacle blocks the path and hence a new one needs to be followed (hence the large peaks and change in direction in the figures when comparing static and dynamic worlds).}
    \label{fig:path_smoothness}
\end{figure}

\begin{table*}[h]
\begin{tabular}{|c|cc|cc|cc|cc|cc|cc|cc|cc|}
\hline
\multirow{2}{*}{} & \multicolumn{2}{c|}{$p_s$}                       & \multicolumn{2}{c|}{$p_{e,x}$}                   & \multicolumn{2}{c|}{$p_{e,y}$}                   & \multicolumn{2}{c|}{$p_{e,z}$}                  & \multicolumn{2}{c|}{$d_{\text{travelled}}$}      & \multicolumn{2}{c|}{$A_{\text{between}}$}         & \multicolumn{2}{c|}{$p_{\text{acc}}$}            & \multicolumn{2}{c|}{$T_{\text{taken}}$}          \\ \cline{2-17} 
                  & \multicolumn{1}{c|}{\textbf{DWB}} & \textbf{TEB} & \multicolumn{1}{c|}{\textbf{DWB}} & \textbf{TEB} & \multicolumn{1}{c|}{\textbf{DWB}} & \textbf{TEB} & \multicolumn{1}{c|}{\textbf{DWB}} & \textbf{TEB} & \multicolumn{1}{c|}{\textbf{DWB}} & \textbf{TEB} & \multicolumn{1}{c|}{\textbf{DWB}} & \textbf{TEB} & \multicolumn{1}{c|}{\textbf{DWB}} & \textbf{TEB} & \multicolumn{1}{c|}{\textbf{DWB}} & \textbf{TEB} \\ \hline

A                 & \multicolumn{1}{c|}{$11.82$}      & $16.67$      & \multicolumn{1}{c|}{$5.36$}       & $6.86$       & \multicolumn{1}{c|}{$7.15$}       & $9.38$       & \multicolumn{1}{c|}{$0.24$}       & $0.32$       & \multicolumn{1}{c|}{$24.82$}      & $25.06$      & \multicolumn{1}{c|}{$1.27$}       & $2.68$       & \multicolumn{1}{c|}{$0.25$}       & $0.14$       & \multicolumn{1}{c|}{$33.88$}      & $44.9$      \\ \hline 
$\text{A}^\prime$ & \multicolumn{1}{c|}{$88.91$}      & $74.14$      & \multicolumn{1}{c|}{$9.12$}       & $10.82$      & \multicolumn{1}{c|}{$10.88$}      & $13.61$      & \multicolumn{1}{c|}{$0.40$}       & $0.50$       & \multicolumn{1}{c|}{$36.64$}      & $34.91$      & \multicolumn{1}{c|}{$1.86$}       & $3.72$       & \multicolumn{1}{c|}{$0.30$}       & $0.14$       & \multicolumn{1}{c|}{$54.96$}      & $68.9$      \\ \hline 
B                 & \multicolumn{1}{c|}{$28.98$}      & $21.37$      & \multicolumn{1}{c|}{$16.35$}      & $33.51$      & \multicolumn{1}{c|}{$11.61$}       & $22.68$      & \multicolumn{1}{c|}{$0.36$}       & $0.50$       & \multicolumn{1}{c|}{$44.06$}      & $44.72$      & \multicolumn{1}{c|}{$2.73$}       & $5.41$       & \multicolumn{1}{c|}{$0.09$}       & $0.05$       & \multicolumn{1}{c|}{$58.97$}      & $80.9$      \\ \hline 
$\text{B}^\prime$                 & \multicolumn{1}{c|}{$18.81$}      & $31.79$      & \multicolumn{1}{c|}{$13.00$}      & $16.99$      & \multicolumn{1}{c|}{$9.70$}       & $13.00$      & \multicolumn{1}{c|}{$0.39$}       & $0.52$       & \multicolumn{1}{c|}{$44.97$}      & $45.95$      & \multicolumn{1}{c|}{$1.98$}       & $5.21$       & \multicolumn{1}{c|}{$0.13$}       & $0.09$       & \multicolumn{1}{c|}{$62.14$}      & $83.5$      \\ \hline 
$\text{B}^{\prime\prime}$ & \multicolumn{1}{c|}{$17.85$}      & $17.54$      & \multicolumn{1}{c|}{$9.70$}       & $11.77$      & \multicolumn{1}{c|}{$9.02$}       & $11.81$      & \multicolumn{1}{c|}{$15.91$}      & $18.04$      & \multicolumn{1}{c|}{$44.93$}      & $44.51$      & \multicolumn{1}{c|}{$1.75$}       & $4.86$       & \multicolumn{1}{c|}{$0.15$}       & $0.10$       & \multicolumn{1}{c|}{$62.01$}      & $78.7$      \\ \hline 
C                 & \multicolumn{1}{c|}{$9.47$}       & $8.69$       & \multicolumn{1}{c|}{$3.17$}       & $4.35$       & \multicolumn{1}{c|}{$4.29$}       & $5.79$       & \multicolumn{1}{c|}{$0.84$}       & $1.15$       & \multicolumn{1}{c|}{$15.00$}      & $15.62$      & \multicolumn{1}{c|}{$1.62$}       & $1.54$       & \multicolumn{1}{c|}{$0.11$}       & $0.18$       & \multicolumn{1}{c|}{$20.54$}      & $28.1$      \\ \hline 
$\text{C}^\prime$ & \multicolumn{1}{c|}{-}            & $84.61$      & \multicolumn{1}{c|}{-}            & $9.85$       & \multicolumn{1}{c|}{-}            & $13.08$      & \multicolumn{1}{c|}{-}            & $2.7$        & \multicolumn{1}{c|}{-}            & $34.46$      & \multicolumn{1}{c|}{-}            & $6.23$       & \multicolumn{1}{c|}{-}            & $0.19$       & \multicolumn{1}{c|}{-}            & $70.5$      \\ \hline 
\end{tabular}
\caption{Summary of the data collected in the simulations. The letters represent specific worlds (A for the playground, B for the office, and C for the warehouse), and primes represent their dynamic versions. The $\text{B}^{\prime\prime}$ row is the data from the simulation where the arm was moved while the base was moving. For $\text{C}^\prime$, the algorithm DWB could not avoid the added obstacle and hence data could not be collected.}
\label{tab:metrics_data}
\end{table*}

The metric that quantifies the load's stability ($\mathbf{p}_e$) can be seen in Fig. \ref{fig:ee_smoothness}. Data on the average performance of algorithms can be found in Table \ref{tab:metrics_data}. We can see that, on average, DWB leads to more stable paths at the end-effector, suggesting that it provides more stable load carrying capabilities.



It is also worth noting how a change in the environment affects the performance of local planners in this regard. When dynamic obstacles are introduced to the simulated environment, the performance of both planners visibly degrades, and the smoothness of the motion of the end-effector suffers. This may be seen by comparing dynamic and static figures in \ref{fig:ee_smoothness} or simply by checking the average smoothness of the end-effector, as in Table \ref{tab:metrics_data}. One reason for this drawback is the fact that the robot would have to make sudden changes to its path to avoid an obstacle; in some cases, it comes to a complete stop before planning a new path. These sudden movements and brakes lead to unplanned and unexpected movement from the robot. This is further amplified by the fact that the local planner only accounts for the acceleration of the mobile base and does not consider the forces applied to the arm of the manipulator. The drivers of the arm motors would work to oppose these movements, leading to higher energy usage than expected. While DWB has a more stable performance than the performance of TEB in dynamic environments, TEB is more effective at avoiding unexpected obstacles than DWB (TEB has less of an increase than DWB when comparing static and dynamic world smoothness metric values.) This can also be seen in the moving arm test scenario ($D^{\prime\prime}$ in table \ref{tab:metrics_data}), where the smoothness values (both path at the base and end-effector) were improved, possibly due to the fact that the arm was moved to a more compact and robust configuration, yet the improvement in smoothness for TEB was still higher than that of DWB. A possible explanation for the increase in the z-axis could be due to the measurement timing mismatch leading to incorrect data being compared. 

Most non-smooth segments in the path are at segments of the path where the robot needs to make large turns, which for static worlds are at the path's beginning or the end. At those points, the robot needs to make significant turns; hence, the angle between each path segment increases by a sizable amount. In some cases, however, that is not true. Even in a static world, the robot can find itself too close to an obstacle and must re-adapt its approach. Such a maneuver will significantly increase the angles between the trajectory segments. This issue would be quite prevalent in more narrow pathways. As it is possible that there may be obstacles in the planned path in a dynamic environment, the robot must make significant turns to follow a newly generated trajectory to avoid the obstacle. This decreases the smoothness of the path and forces the robot to approach the desired end pose from an unfavorable direction in some cases, which can also decrease the overall smoothness of the path (see how the addition of obstacles in Fig. \ref{fig:path_smoothness} creates a large peak in the angles between directions.) 

\begin{figure}[H]
    \centering
        \includegraphics[width=0.8\linewidth]{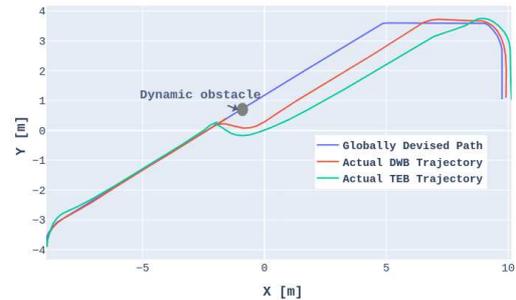}
        \label{fig:playground_dynamic}
        \caption{Trajectories of the mobile manipulator in the playground environment with a single obstacle in the path. The global planner has devised a path which the robot initially is following until it encounters the obstacle. This figure demonstrates how both local planners react in such situations. As it can be seen, TEB reacts more radically at the beginning, but soon stabilizes on a straight trajectory, while DWB tries to get the robot back to the globally devised trajectory.}
    \label{fig:playgound_trajectory}
\end{figure}

The results for accuracy and total time in general can be summarized as follows: 
DWB has faster performance than TEB, and TEB has a more accurate final position than DWB. We can also see that TEB tends to drive further, except when there are obstacles in the way, in which case TEB would travel less. 

We can see that TEB, on average, deviates more from the global path than DWB, as suggested by the $A_{\text{between}}$ values. DWB more easily follows the global path. However, the increase in the area between the global paths and the local trajectory caused by unforeseen obstacles is smaller for TEB than for DWB ($\approx 39 \%$ increase for TEB and $\approx 47 \%$ increase for DWB), see \figref{fig:playgound_trajectory}.


It is important to mention scenarios in which specific local planners failed to find/execute a path to the goal. In the world of wide mazes, there is a tight corridor in the map through which the robot would have to travel to reach the goal. The scenario in which DWB failed was dynamic warehouse. The hypothesised reason for this failure is attributed to the size of the added obstacles. As the sampling window which the DWB algorithm uses to find viable paths is small, an obstacle that covers the window would lead to the robot being unable to find a path. In most cases, this happens due to the poor reaction of the algorithm where the robot drives too close to the obstacle, leading to the covering of sample window. If this problem is noticed soon enough, it can be avoided by calling the global planner and re-planning a new, viable plan around the new obstacle. Finally, seeing how for most metrics, the increase in value was less for TEB than for DWB, we can conclude that while TEB was affected less by the obstacles, DWB still has a more reliable and smooth performance in both static and dynamic environments. The data and code to validate these conclusions can be found in our \footnote{\url{https://github.com/sevag-tafnakaji/local_planner_benchmarking}}{GitHub page.} 

\section{CONCLUSION}
\label{section:conclusion}

In this paper, we present a set of measures to benchmark the performance of local motion planners for mobile manipulator navigation and a software framework to implement these benchmarks conveniently. The stability of the manipulator is one of the most critical considerations in planning the trajectory of a mobile manipulator. The robot should move without any sudden stop, violent twist, or pull of the load. In this paper, we quantify the motion's smoothness for the load being carried by a mobile manipulator. We benchmark two prominent local planners, DWB and TEB, to demonstrate which one delivers the most smooth trajectory. The findings of this paper indicate that, on average, DWB results in a trajectory that causes less disturbance to the manipulator in comparison to TEB. Nonetheless, when an unforeseen obstacle appears in the scope of the robot's sensors on its way to the goal, TEB can handle the transition to the exogenous path around the obstacle more efficiently than its DWB counterpart, however DWB proved to still provide smoother paths (both for the end-effector and the mobile base). We plan to extend this work by utilizing these measures to find the safest trajectories for a real mobile manipulator in order to achieve various tasks successfully. 

\section*{ACKNOWLEDGMENT}
This work was supported by by Chalmers AI Research Center (CHAIR), Chalmers Gender Initiative for Excellence (Genie), and the project AIMCoR - AI-enhanced Mobile Manipulation Robot for Core Industrial Applications.

\bibliographystyle{IEEEtran}

\bibliography{main.bib}

\begin{thebibliography}{10}
\providecommand{\url}[1]{#1}
\csname url@samestyle\endcsname
\providecommand{\newblock}{\relax}
\providecommand{\bibinfo}[2]{#2}
\providecommand{\BIBentrySTDinterwordspacing}{\spaceskip=0pt\relax}
\providecommand{\BIBentryALTinterwordstretchfactor}{4}
\providecommand{\BIBentryALTinterwordspacing}{\spaceskip=\fontdimen2\font plus
\BIBentryALTinterwordstretchfactor\fontdimen3\font minus
  \fontdimen4\font\relax}
\providecommand{\BIBforeignlanguage}[2]{{%
\expandafter\ifx\csname l@#1\endcsname\relax
\typeout{** WARNING: IEEEtran.bst: No hyphenation pattern has been}%
\typeout{** loaded for the language `#1'. Using the pattern for}%
\typeout{** the default language instead.}%
\else
\language=\csname l@#1\endcsname
\fi
#2}}
\providecommand{\BIBdecl}{\relax}
\BIBdecl

\bibitem{marder-eppstein_office_2010}
E.~Marder-Eppstein, E.~Berger, T.~Foote, B.~Gerkey, and K.~Konolige, ``The
  office marathon: Robust navigation in an indoor office environment,'' in
  \emph{2010 {IEEE} International Conference on Robotics and Automation}, 2010,
  pp. 300--307, {ISSN}: 1050-4729.

\bibitem{hidalgo-paniagua_solving_2017}
\BIBentryALTinterwordspacing
A.~Hidalgo-Paniagua, M.~A. Vega-Rodríguez, J.~Ferruz, and N.~Pavón, ``Solving
  the multi-objective path planning problem in mobile robotics with a
  firefly-based approach,'' \emph{Soft Computing}, vol.~21, no.~4, pp.
  949--964, 2017. [Online]. Available:
  \url{http://link.springer.com/10.1007/s00500-015-1825-z}
\BIBentrySTDinterwordspacing

\bibitem{zhang_multilevel_2019}
X.~Zhang, J.~Wang, Y.~Fang, and J.~Yuan, ``Multilevel humanlike motion planning
  for mobile robots in complex indoor environments,'' \emph{{IEEE} Transactions
  on Automation Science and Engineering}, vol.~16, no.~3, pp. 1244--1258, 2019,
  conference Name: {IEEE} Transactions on Automation Science and Engineering.

\bibitem{wang_eb-rrt_2020}
J.~Wang, M.~Q.-H. Meng, and O.~Khatib, ``{EB}-{RRT}: Optimal motion planning
  for mobile robots,'' \emph{{IEEE} Transactions on Automation Science and
  Engineering}, vol.~17, no.~4, pp. 2063--2073, 2020, conference Name: {IEEE}
  Transactions on Automation Science and Engineering.

\bibitem{kudriashov_introduction_2020}
\BIBentryALTinterwordspacing
A.~Kudriashov, T.~Buratowski, M.~Giergiel, and P.~Małka, \emph{Introduction to
  Mobile Robots Navigation, Localization and Mapping}.\hskip 1em plus 0.5em
  minus 0.4em\relax Springer International Publishing, 2020, vol.~87, pp.
  7--38, series Title: Mechanisms and Machine Science. [Online]. Available:
  \url{http://link.springer.com/10.1007/978-3-030-48981-6_2}
\BIBentrySTDinterwordspacing

\bibitem{fox_dynamic_1997}
D.~Fox, W.~Burgard, and S.~Thrun, ``The dynamic window approach to collision
  avoidance,'' vol.~4, no.~1, pp. 23--33, conference Name: {IEEE} Robotics \&
  Automation Magazine.

\bibitem{rosmann_integrated_2017}
\BIBentryALTinterwordspacing
C.~Rösmann, F.~Hoffmann, and T.~Bertram, ``Integrated online trajectory
  planning and optimization in distinctive topologies,'' \emph{Robotics and
  Autonomous Systems}, vol.~88, pp. 142--153, 2017. [Online]. Available:
  \url{https://www.sciencedirect.com/science/article/pii/S0921889016300495}
\BIBentrySTDinterwordspacing

\bibitem{roesmann_trajectory_2012}
C.~Roesmann, W.~Feiten, T.~Woesch, F.~Hoffmann, and T.~Bertram, ``Trajectory
  modification considering dynamic constraints of autonomous robots,'' in
  \emph{{ROBOTIK} 2012; 7th German Conference on Robotics}, pp. 1--6.

\bibitem{Quinlan1993}
S.~Quinlan and O.~Khatib, ``Elastic bands: connecting path planning and
  control,'' in \emph{[1993] Proceedings IEEE International Conference on
  Robotics and Automation}, 1993, pp. 802--807 vol.2.

\bibitem{pittner_systematic_2018}
M.~Pittner, M.~Hiller, F.~Particke, L.~Patino-Studencki, and J.~Thielecke,
  ``Systematic analysis of global and local planners for optimal trajectory
  planning,'' in \emph{{ISR} 2018; 50th International Symposium on Robotics},
  2018, pp. 1--4.

\bibitem{cybulski_accuracy_2019}
B.~Cybulski, A.~Wegierska, and G.~Granosik, ``Accuracy comparison of navigation
  local planners on {ROS}-based mobile robot,'' in \emph{2019 12th
  International Workshop on Robot Motion and Control ({RoMoCo})}, 2019, pp.
  104--111, {ISSN}: 2575-5579.

\bibitem{marin-plaza_global_2018}
\BIBentryALTinterwordspacing
P.~Marin-Plaza, A.~Hussein, D.~Martin, and A.~d.~l. Escalera, ``Global and
  local path planning study in a {ROS}-based research platform for autonomous
  vehicles,'' \emph{Journal of Advanced Transportation}, vol. 2018, pp. 1--10,
  2018. [Online]. Available:
  \url{https://www.hindawi.com/journals/jat/2018/6392697/}
\BIBentrySTDinterwordspacing

\bibitem{naotunna_comparison_2020}
I.~Naotunna and T.~Wongratanaphisan, ``Comparison of {ROS} local planners with
  differential drive heavy robotic system,'' in \emph{2020 International
  Conference on Advanced Mechatronic Systems ({ICAMechS})}, 2020, pp. 1--6,
  {ISSN}: 2325-0690.

\bibitem{hsieh_experimental_2016}
\BIBentryALTinterwordspacing
C.~Sprunk, J.~Röwekämper, G.~Parent, L.~Spinello, G.~D. Tipaldi, W.~Burgard,
  and M.~Jalobeanu, ``An experimental protocol for benchmarking robotic indoor
  navigation,'' in \emph{Experimental Robotics}, M.~A. Hsieh, O.~Khatib, and
  V.~Kumar, Eds.\hskip 1em plus 0.5em minus 0.4em\relax Springer International
  Publishing, 2016, vol. 109, pp. 487--504, series Title: Springer Tracts in
  Advanced Robotics. [Online]. Available:
  \url{http://link.springer.com/10.1007/978-3-319-23778-7_32}
\BIBentrySTDinterwordspacing

\bibitem{wen_mrpb_2021}
\BIBentryALTinterwordspacing
J.~Wen, X.~Zhang, Q.~Bi, Z.~Pan, Y.~Feng, J.~Yuan, and Y.~Fang, ``{MRPB} 1.0: A
  unified benchmark for the evaluation of mobile robot local planning
  approaches,'' number: {arXiv}:2011.00491. [Online]. Available:
  \url{http://arxiv.org/abs/2011.00491}
\BIBentrySTDinterwordspacing

\bibitem{guillen_ruiz_measuring_2020}
\BIBentryALTinterwordspacing
S.~Guillén~Ruiz, L.~V. Calderita, A.~Hidalgo-Paniagua, and J.~P.
  Bandera~Rubio, ``Measuring smoothness as a factor for efficient and socially
  accepted robot motion,'' \emph{Sensors}, vol.~20, no.~23, p. 6822, 2020,
  number: 23 Publisher: Multidisciplinary Digital Publishing Institute.
  [Online]. Available: \url{https://www.mdpi.com/1424-8220/20/23/6822}
\BIBentrySTDinterwordspacing

\end{thebibliography}

\end{document}